\documentclass[runningheads]{llncs}
\usepackage[colorlinks,
            linkcolor=blue,
            anchorcolor=blue,
            urlcolor=blue,
            citecolor=blue]{hyperref}
\usepackage{graphicx}
\usepackage{bbding}
\usepackage{makecell}
\usepackage{enumitem}
\usepackage{amsmath}
\usepackage{amsfonts}
\usepackage{multirow}
\usepackage{array}
\usepackage{bm}
\usepackage[misc]{ifsym}

\begin{document}
\title{Lesion-based Contrastive Learning for Diabetic Retinopathy Grading from Fundus Images}
\titlerunning{Lesion-based Contrastive Learning}
%

\author{Yijin Huang$^{1}$, Li Lin$^{1,2}$, Pujin Cheng$^{1}$, Junyan Lyu$^{1}$, Xiaoying Tang$^{1(\textrm{\Letter)}}$}

\authorrunning{Y. Huang et al.}

%
\institute{Department of Electrical and Electronic Engineering, \\ Southern University of Science and Technology, Shenzhen, China \\ \email{tangxy@sustech.edu.cn} \and
School of Electronics and Information Technology, \\ Sun Yat-sen University, Guangzhou, China
}
\maketitle              
\begin{abstract}
Manually annotating medical images is extremely expensive, especially for large-scale datasets. Self-supervised contrastive learning has been explored to learn feature representations from unlabeled images. However, unlike natural images, the application of contrastive learning to medical images is relatively limited. In this work, we propose a self-supervised framework, namely lesion-based contrastive learning for automated diabetic retinopathy (DR) grading. Instead of taking entire images as the input in the common contrastive learning scheme, lesion patches are employed to encourage the feature extractor to learn representations that are highly discriminative for DR grading. We also investigate different data augmentation operations in defining our contrastive prediction task. Extensive experiments are conducted on the publicly-accessible dataset EyePACS, demonstrating that our proposed framework performs outstandingly on DR grading in terms of both linear evaluation and transfer capacity evaluation.

\end{abstract}
\section{Introduction}
\begin{figure}[t]
    \centering
    \includegraphics[width=8cm]{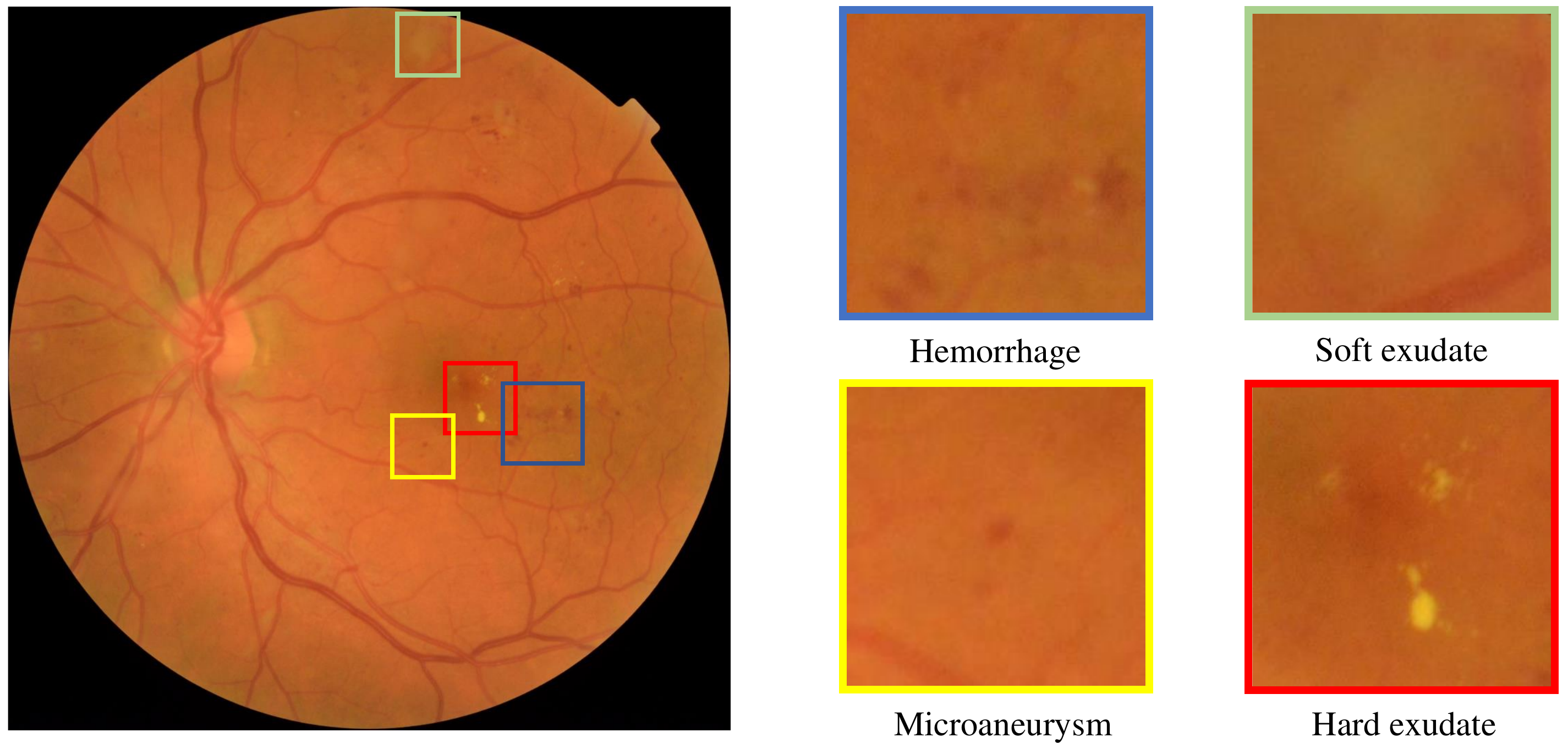}
    \caption{A representative fundus image with four types of DR related lesions.}
    \label{fig:fundus}
\end{figure}

Diabetic retinopathy (DR) is one of the microvascular complications of diabetes, which may cause vision impairments and even blindness \cite{ref_proc1}. The major pathological signs of DR include hemorrhages, exudates, microaneurysms, and retinal neovascularization, as shown in Fig. \ref{fig:fundus}. The color fundus image is the most widely-used photography for ophthalmologists to identify the severity of DR, which can clearly reveal the presence of different lesions. Early diagnoses and timely interventions are of vital importance in preventing DR patients from vision malfunction. As such, automated and efficient fundus image based DR diagnosis systems are urgently needed.

Recently, deep learning has achieved great success in the field of medical image analysis. Convolutional neural networks (CNNs) have been proposed to tackle the automated DR grading task \cite{cab,AFN,zoom}. The success of CNN is mainly attributed to its capability of extracting highly representative features. However, it usually requires a large-scale annotated dataset to train a network. The annotation process is time-intensive, tedious, and error-prone, and hence ophthalmologists bear a heavy burden in building a well-annotated dataset.

Self-supervised learning (SSL) methods \cite{rotate,sequence,mri,cube} have been explored to learn feature representations from unlabeled images. As a representative SSL method, contrastive learning (CL) \cite{bach,simclr,ultra} has been very successful in the natural image field, which defines a contrastive prediction task that trying maximize the similarity between features from differently augmented views of the same image and simultaneously maximize the distance between features from different images, via a contrastive loss. However, the application of CL in the fundus image field \cite{mulm,sia} is relatively limited, mainly due to fundus images' high resolution and low proportion of diagnostic features. First, fundus images are of high resolution so as to clearly reveal structural and pathological details. And it is challenging for CL to train high-resolution images with large batch sizes which are nevertheless generally required by CL so as to provide more negative samples. Secondly, data augmentation is typically applied in CL to generate different views of the same image. However, some strong data augmentation operations applied to fundus images may destroy important domain-specific information, such as cropping and Cutout \cite{cutout}. In a natural image, salient objects generally occupy a very large proportion and each part of them may contribute to the recognition of the object of interest, whereas the diagnostic features in a fundus image, such as lesions, may only occupy a small part of the whole image. This may result in that most cropped patches inputted to the CL framework have few or no diagnostic features. In this way, the network is prone to learn feature representations that are distinguishable for different views of the image but not discriminative for downstream tasks. CL in fundus image based DR grading is even rarer.

To address the aforementioned issues, we propose a lesion-based contrastive learning approach for fundus image based DR grading. Instead of using entire fundus images, lesion patches are taken as the input for our contrastive prediction task. By focusing on  patches with lesions, the network is encouraged to learn more discriminative features for DR grading. The main steps of our framework are as follows. First, an object detection network is trained on a publicly-available dataset IDRiD \cite{IDRiD} that consists of 81 fundus images with annotations of lesions. Then, the detection network is applied to the training set of EyePACS \cite{EyePACS} to predict lesions with a relatively high confidence threshold. Next, random data augmentation operations are applied to the lesion patches to generate multiple views of them. The feature extraction network in our CL framework is expected to map inputted patches into an embedding feature space, wherein the similarity between features from different views of the same lesion patch and the distance between features from different patches are maximized, by minimizing a contrastive loss. The performance of our proposed method is evaluated based on linear evaluation and transfer capacity evaluation on EyePACS.

The main contributions of this paper are three-fold: (1) We present a self-supervised framework for DR grading, namely lesion-based contrastive learning. Our framework's contrastive prediction task takes lesion patches as the input, which addresses the problem of high memory requirements and lacking diagnostic features, as to common CL schemes. This design can be easily extended to other types of medical images with relatively weak physiological characteristics. (2) We study different data augmentation operations in defining our contrastive prediction task. Results show that a composition of cropping, color distortion, gray scaling and rotation is beneficial in defining pretext tasks for fundus images to learn discriminative feature representations. (3) We evaluate our framework on the large-scale EyePACS dataset for DR grading. Results from linear evaluation and transfer capacity evaluation identify our method's superiority. The source code is available at \href{https://github.com/YijinHuang/Lesion-based-Contrastive-Learning}{https://github.com/YijinHuang/Lesion-based-Contrastive-Learning}

\begin{figure}[t]
    \centering
    \includegraphics[width=\linewidth]{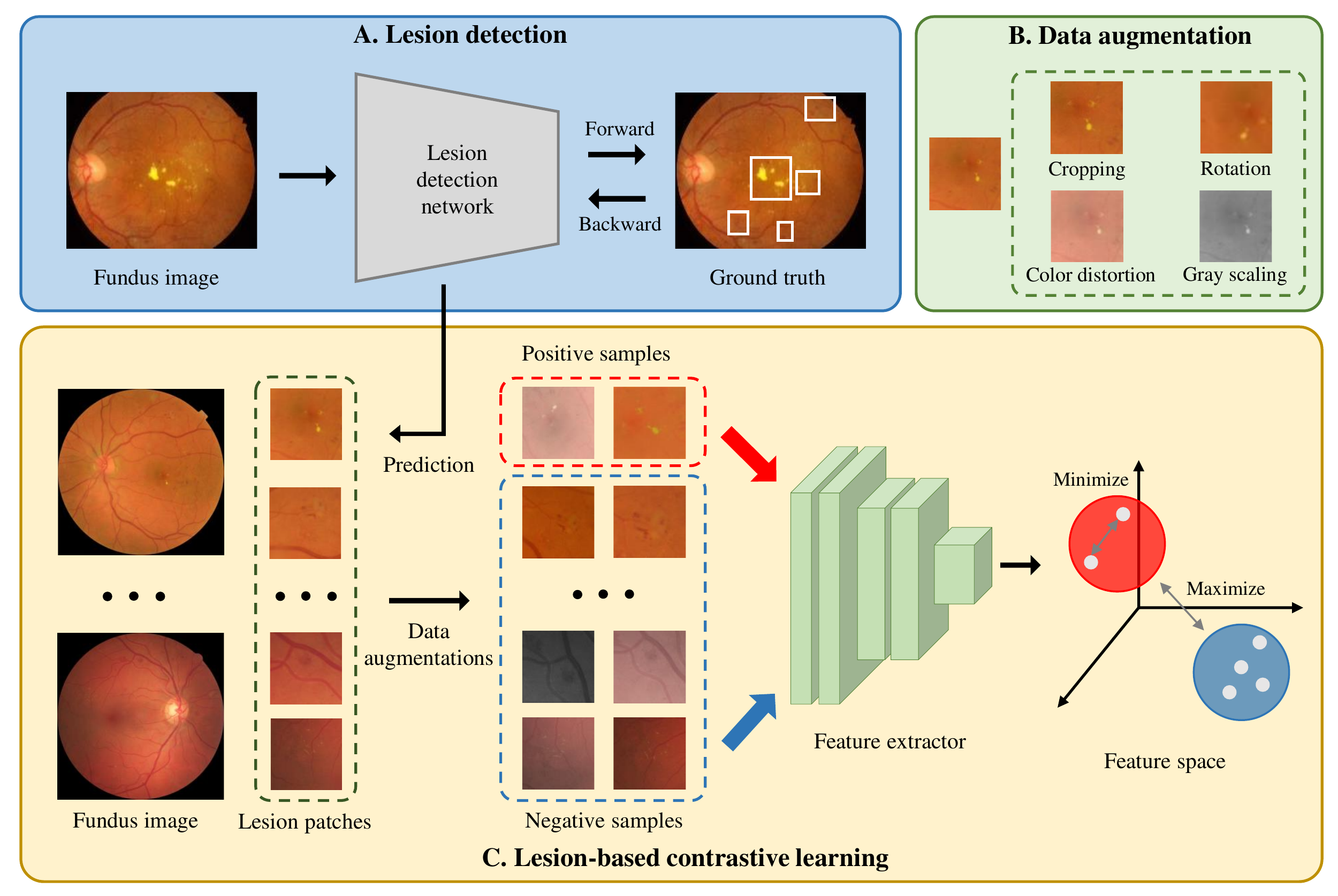}
    \caption{The proposed framework. In part A, an object detection network is trained to predict lesion patches in fundus images for subsequent contrastive learning. Illustrations of data augmentation operations are provided in part B. In part C, lesion patches are processed by a composition of data augmentation operations to generate multiple views of the lesion patches. A feature extractor is applied to encode patches into an embedding feature space that minimizes a contrastive loss.}
    \label{fig:framework}
\end{figure}

\section{Methods}

\subsection{Generation of lesion patches}
Two datasets are used in this work. An object detection network is trained on one dataset with lesion annotations. Then, this detection network is used to generate lesion patches of fundus images from the other dataset for subsequent contrastive learning. Because of limited training samples in the first dataset, the detection network has a relatively poor generalization ability and cannot precisely predict lesions of fundus images from the other dataset. Therefore, a high confidence threshold is set to eliminate unconfident predictions. Then, we resize all fundus images to $512 \times 512$, and the bounding boxes of the patches are scaled correspondingly. After that, we expand every predicted bounding box to $128 \times 128$ with the lesion lying in the center and then randomly shift the box within a range such that the resulting box still covers the lesion. In this way, we increase the difficulty of the contrastive prediction task while ensure the training of our CL framework can be performed with a large batch size. Please note the lesion detection network is not involved in the testing phase.

\subsection{Generation of multiple views of lesion patches} \label{lesionpatches}
Rather than employing a carefully designed task to learn feature representations \cite{rotate,sequence}, CL tries to maximize the agreement between differently augmented views of the same image. Data augmentation, a widely used technique in deep learning, is applied to generate different views of a sample in CL. Some data augmentation operations that are commonly used in the natural image field may destroy important diagnostic features when transferred to the fundus image field. Therefore, as shown in part B of Fig. \ref{fig:framework}, four augmentation operations are considered in our work: cropping, color distortion, gray scaling, and rotation.

Let $D = \{x_i, i=1, ..., N\}$ denote a randomly-sampled batch with a batch size of $N$. Two random compositions of data augmentation operators $(\mathcal{T}, \mathcal{T}^\prime)$ are applied to each data point $x_i$ to generate two different views $(\mathcal{T}(x_i), \mathcal{T}^\prime(x_i))$. Note that the parameters of these augmentation operators may differ from data point to data point. Now, we obtain a new batch $D^\prime = \{\tilde{x}_i, i=1, ..., 2N\}$. Given a patch $\tilde{x}_i$ that is generated from $x_i$, we consider $\tilde{x}_{ j\vert j \neq i}$ that is also generated from $x_i$ as a positive sample $\tilde{x}_i^+$ and every patch in the set $\Lambda_i^- = \{\tilde{x}_k\}_{k \neq i, k \neq j}$ as a negative sample $\tilde{x}_i^-$.

\subsection{Lesion-based contrastive learning} \label{lcl}
Given a data point $\tilde{x}_i$, we first use a feature extractor $f(\cdot)$ to extract its feature vector $h_i$. Specificly, $f(\cdot)$ is a CNN and $h_i$ is the feature vector right before the fully connected layer of the network. Then, a projection head $g(\cdot)$ is applied to map the feature vector into an embedding space to obtain $z_i$. Given a batch $D^\prime$, we define $Z = \{z_i | z_i = g(f(\tilde{x}_i)), \tilde{x}_i \in D^\prime\}$ and $\Omega^- = \{z_i^- | z_i^- = g(f(\tilde{x}_i^-)), {\tilde{x}_i^- \in \Lambda_i^-}\}$. For every $z_i$, our contrastive prediction task is to identify embedded feature $z_i^+ = g(f(\tilde{x}_i^+))$ from $Z$. To find $z_i^+$, we define the one having the highest cosine similarity with $z_i$ as our prediction. To maximize the similarity of positive samples and to minimize that of negative samples, we define our contrastive loss as
\begin{equation}
    \mathcal{L}(Z) = -\sum\limits_{i=1}^{2N} \log \frac{\exp(\text{sim}(\bm{z}_i, \bm{z}_i^+) / \tau)}{\exp(\text{sim}(\bm{z}_i, \bm{z}_i^+) / \tau) + \sum_{z_i^- \in \Omega^-}\exp(\text{sim}(\bm{z}_i, \bm{z}_i^-) / \tau)},
\end{equation}
where $N$ denotes the batch size of $Z$, $\text{sim}(\bm{z}_i, \bm{z}_j) = \bm{z}_i^T\bm{z}_j / \lVert\bm{z}_i\rVert\lVert\bm{z}_j\rVert$, and $\tau$ is a temperature parameter. In the testing phase, we do not use the projection head $g(\cdot)$ but only the feature extractor $f(\cdot)$ for downstream tasks. Our framework of lesion-based CL is depicted in part C of Fig. \ref{fig:framework}.

\subsection{Implementation details}

\subsubsection{Data augmentation operations.} For the cropping operation, we randomly crop lesion patches with a random factor in [0.8, 1.2]. For the gray scaling operation, each patch has a 0.2 probability of being gray scaled. The color distortion operation adjusts the brightness, contrast, and saturation of patches with a random factor in [-0.4, 0.4] and also changes the hue with a random factor in [-0.1, 0.1]. The rotation operation randomly rotates patches by an arbitrary angle.

\subsubsection{Lesion detection network.} Faster-RCNN \cite{faster} with ResNet50 \cite{resnet} as the backbone is adopted as our lesion detection network. We apply transfer learning by initializing the network with parameters from a model pre-trained on the COCO \cite{coco} dataset. The detection network is trained with Adam optimizer for 100 epochs, with a 0.01 initial learning rate and getting decayed by 0.1 at the 50th epoch and the 80th epoch.

\subsubsection{Contrastive learning network.} We also use ResNet50 as our feature extractor. The projection head is a one-layer MLP with ReLU as the activation function, which reduces the dimension of the feature vector to $128$. We adopt SGD optimizer with a 0.001 initial learning rate and cosine decay strategy to train the network. The batch size is set to be 768 and the temperature parameter $\tau$ is set to be $0.07$. The augmented views of lesion patches are resized to $128 \times 128$ as the input to our contrastive learning task. All experiments are equally trained for 1000 epochs with a fixed random seed.

\begin{table}[h]
    \centering
    \caption{The total number of lesion patches for the lesion-based CL under different confidence thresholds of the lesion detection network with a 31.17\% mAP.}
    \setlength{\tabcolsep}{7pt}
    \begin{tabular}{cccccccc}  \hline
        Confidence threshold & \# images & \# lesions  \\ 	\hline
        0.7 & 25226 & 88867 \\
        0.8 & 21578 & 63362 \\
        0.9 & 15889 & 35550 \\
        \hline
    \end{tabular}
    \label{table1}
\end{table}
\subsection{Evaluation protocol}
\subsubsection{Linear evaluation.} Linear evaluation is a widely used method for evaluating the quality of the learned representations of a self-supervised model. The pre-trained feature extractor described in Section \ref{lcl} is frozen and a linear classifier on top of it is trained in a fully-supervised manner. The performance of downstream tasks is then used as a proxy of the quality of the learned representations.

\subsubsection{Transfer capacity evaluation.} Pre-trained parameters of the feature extractor can be transferred to models used in downstream tasks. To evaluate the transfer learning capacity, we unfreeze and fine-tune the feature extractor followed by a linear classifier in supervised downstream tasks.

\section{Experiment}

\subsection{Dataset and evaluation metric}
\subsubsection{IDRiD.} IDRiD \cite{IDRiD} consists of 81 fundus images, with pixel-wise lesion annotations of hemorrhages, microaneurysms, soft exudates, and hard exudates. These manual annotations are converted to bounding boxes to train an object detection network \cite{hem}. Microaneurysms are excluded in this work because detecting them is challenging and will lead to a large number of false positive predictions. In training the lesion detection network, 54 samples are used for training and 27 for validation.

\subsubsection{EyePACS.} 35k/11k/43k fundus images are provided in EyePACS \cite{EyePACS} for training/validation/testing (the class distribution of EyePACS is shown in Fig. \url{A1} of the appendix). According to the severity of DR, images are classified into five grades: 0 (normal), 1 (mild), 2 (moderate), 3 (severe), and 4 (proliferative). The training and validation sets without annotations are used for training our self-supervised model. The total number of lesion patches under different confidence thresholds of the detection network is shown in Table \ref{table1}. Representative lesion detection results are provided in Fig. \url{A2} of the appendix. Partial datasets are obtained by randomly sampling 1\%/5\%/10\%/25\%/100\% (0.3k/1.7k/3.5k/8.7k/35k images) from the training set, together with the corresponding annotations. Images in the partial datasets and the test set are resized to $512 \times 512$ for training and testing subsequent DR grading models in both linear evaluation and transfer capability evaluation settings.

\subsubsection{Evaluation metric.} We adopt the quadratic weighted kappa \cite{EyePACS} for evaluation, which works well for unbalanced datasets.

\subsection{Composition of data augmentation operations}
We evaluate the importance of an augmentation operation by removing it from the composition or applying it solely. As shown in the top panel of Table \ref{table2}, it is insufficient for a single augmentation operation to learn discriminative representations. Even so, color distortion works much better than other operations, showing its importance in defining our contrastive prediction task for fundus images. This clearly indicates that DR grading benefits from color invariance. We conjecture it is because the EyePACS images are highly diverse, especially in terms of image intensity profiles. From the bottom panel of Table \ref{table2}, we notice that an absence of any of the four augmentation operations leads to a decrease in performance. Applying the composition of all four augmentation operations considerably increases the difficulty of our contrastive prediction task, but it also significantly improves the quality of the learned representations. Notably, this contrasts common CL methods; a heavy data augmentation would typically hurt the performance of disease diagnosis \cite{mulm}.

\begin{table}[h]
    \centering
    \caption{Impact of different compositions of data augmentation operations. The kappa is the result of linear evaluation on the 25\% partial dataset, under a detection confidence threshold of 0.8.}
    \setlength{\tabcolsep}{7pt}
    \begin{tabular}{m{2cm}<{\centering}ccccccc}  \hline
        Cropping & Rotation & Color distortion & Gray Scaling & Kappa \\ 	\hline
        \checkmark & & & & 44.25 \\
        & \checkmark & & & 41.71 \\
        & & \checkmark & & \textbf{53.83} \\
        & & & \checkmark & 48.17 \\
        \hline
        & \checkmark & \checkmark & \checkmark & 62.05 \\
        \checkmark &  & \checkmark & \checkmark & 57.94 \\
        \checkmark & \checkmark & & \checkmark & 59.96 \\
        \checkmark & \checkmark & \checkmark & & 61.55 \\
        \checkmark & \checkmark & \checkmark & \checkmark & \textbf{62.49} \\
        \hline
    \end{tabular}
    \label{table2}
\end{table}

\begin{table}[t]
    \caption{Linear evaluation and transfer capacity evaluation results on partial datasets.}
    \centering
    \begin{tabular}{lcccccc}
        \hline
        \multirow{3}{*}{\textcolor[rgb]{0.2,0.2,0.2}{Method} } & \multirow{3}{*}{\textcolor[rgb]{0.2,0.2,0.2}{Confidence }\textcolor[rgb]{0.2,0.2,0.2}{}\textcolor[rgb]{0.2,0.2,0.2}{threshold} } & \multicolumn{5}{c}{\textcolor[rgb]{0.2,0.2,0.2}{Partial dataset} }                                          \\
        \cline{3-7}
        &                                                                                                                                 & 1\%            & 5\%            & 10\%           & 25\%           & 100\%                                  \\
        \cline{3-7}
        &                                                                                                                                 & \multicolumn{5}{c}{\textcolor[rgb]{0.2,0.2,0.2}{Quadratic }\textcolor[rgb]{0.2,0.2,0.2}{weighted kappa} }  \\
        \hline
        \multicolumn{7}{c}{\textcolor[rgb]{0.2,0.2,0.2}{\textit{Linear evaluation}} }                                                                                                                                                                                                                         \\
        \hline
        \textcolor[rgb]{0.2,0.2,0.2}{SimCLR (128 $\times$ 128)}                   & -                                                                                                                               & 16.19              & 26.70              & 31.62              & 37.41              & 43.64                                      \\
        \textcolor[rgb]{0.2,0.2,0.2}{SimCLR (224 $\times$ 224)}                   & -                                                                                                                               & 12.15              & 26.56              & 29.94              & 37.86              & 55.32                                      \\
        \textcolor[rgb]{0.2,0.2,0.2}{Lesion-base CL (ours)}    & 0.7                                                                                                                             & 24.72              & 43.98              & 53.33              & 61.55              & 66.87                                      \\
        \textcolor[rgb]{0.2,0.2,0.2}{Lesion-base CL (ours)}    & 0.8                                                                                                                             & 26.22              & 48.18              & 56.30              & 62.49              & 66.80                                      \\
        \textcolor[rgb]{0.2,0.2,0.2}{Lesion-base CL (ours)}    & 0.9                                                                                                                             & \textbf{31.92}              & \textbf{56.57}              & \textbf{60.70}              & \textbf{63.45}              & \textbf{66.88}                                      \\
        \hline
        \multicolumn{7}{c}{\textcolor[rgb]{0.2,0.2,0.2}{\textit{Transfer capacity evaluation}} }                                                                                                                                                                                                                                \\
        \hline
        \textcolor[rgb]{0.2,0.2,0.2}{Supervised}               & -                                                                                                                               & 53.36              & 72.19              & 76.35              & 79.85              & 83.10                                      \\
        \textcolor[rgb]{0.2,0.2,0.2}{SimCLR (128 $\times$ 128)}                   & -                                                                                                                               & 63.16              & 72.30              & 76.33              & 79.59              & 82.72                                      \\
        \textcolor[rgb]{0.2,0.2,0.2}{SimCLR (224 $\times$ 224)}                   & -                                                                                                                               & 55.43              & 70.66              & 75.15              & 77.32              & 82.11                                      \\
        \textcolor[rgb]{0.2,0.2,0.2}{Lesion-base CL (ours)}    & 0.7                                                                                                                             & \textbf{68.37} & \textbf{75.40} & 77.34          & 80.34          & 82.80                                      \\
        \textcolor[rgb]{0.2,0.2,0.2}{Lesion-base CL (ours)}    & 0.8                                                                                                                             & 68.14          & 74.18          & \textbf{77.49} & \textbf{80.74} & \textbf{83.22}                                      \\
        \textcolor[rgb]{0.2,0.2,0.2}{Lesion-base CL (ours)}    & 0.9                                                                                                                             & 66.43          & 73.85          & 76.93          & 80.27          & 83.04                                      \\
        \hline
    \end{tabular}
    \label{table3}
\end{table}

\subsection{Evaluation on DR grading}
To evaluate the quality of the learned representations from fundus images, we perform linear evaluation and transfer capacity evaluation on the partial datasets. A state-of-the-art CL method SimCLR \cite{simclr} is adopted as our baseline for comparison, which nevertheless takes an entire fundus image as the input for the contrastive prediction task. In Table \ref{table3}, SimCLR (128 $\times$ 128) denotes the multiple views of fundus images are resized to 128 $\times$ 128 as the input for the contrastive prediction task in SimCLR, being consistent with the input size of our lesion-based CL framework. However, crop-and-resize transformation may critically change the pixel size of the input. SimCLR (224 $\times$ 224) experiments are conducted based on the consideration that aligning the pixel size of the input for CL and that for downstream tasks may achieve better performance. The ResNet50 model in the fully-supervised DR grading is initialized with parameters from a model trained on the ImageNet \cite{imgnet} dataset. Training curves are provided in Fig. \url{A3} of the appendix.

As shown in Table \ref{table3}, our lesion-based CL under a detection confidence threshold of 0.9 achieves 66.88\% kappa on linear evaluation on the full training set. Significant improvements over SimCLR have been observed; 23.34\% over SimCLR (128 $\times$ 128) and 11.56\% over SimCLR (224 $\times$ 224). The superiority of our method is more evident when a smaller training set is used for linear evaluation. Note that although using a higher confidence threshold results in fewer lesion patches for training, a better performance is achieved on linear evaluation. This implies that by improving the quality of lesions in the training set of the contrastive prediction task, the model can more effectively learn discriminative feature representations. For the transfer capacity evaluation, when fine-tuning on the full training set, there is not much difference between the fully-supervised method and CL methods. This is because feature representations can be sufficiently learned under full supervison, and thus there may be no need for CL based learning. Therefore, the advantage of our proposed method becomes more evident when the training samples for fine-tuning become fewer. With only 1\% partial dataset for fine-tunning, the proposed method under a confidence threshold of 0.7 has a higher kappa by 15.01\% than the fully-supervised method. Both linear evaluation and transfer capacity evaluation suggest that our proposed method can better learn feature representations, and thus is able to enhace DR grading, by exploiting unlabeled images (see also Fig. \url{A4} in the appendix).

\section{Conclusion}
In this paper, we propose a novel self-supervised framework for DR grading. We use lesion patches as the input for our contrastive prediction task rather than entire fundus images, which encourages our feature extractor to learn representations of diagnostic features, improving the transfer capacity for downstream tasks. We also present the importance of different data augmentation operations in our CL task. By performing linear evaluation and transfer capacity evaluation on EyePACS, we show that our method has a superior DR grading performance, especially when the sample size of the training data with annotations is limited. This work, to the best of our knowledge, is the first one of its kind that has attempted contrastive learning on fundus image based DR grading.

%
%
%
%

\end{document}